\documentclass{llncs}

\usepackage[T2A]{fontenc}
\usepackage[utf8]{inputenc}
\usepackage[english]{babel}
\usepackage{listings}
\begin{document}

\title{Gibberish Semantics: How Good is Russian Twitter in Word Semantic Similarity Task?}
\titlerunning{Gibberish Semantics}  
%
\author{Nikolay N. Vasiliev}
\authorrunning{Nikolay N. Vasiliev} 
%
%
\institute{SpazioDati S.r.l, \\Trento, Viale A.Olivetti, 13, Italy,\\
\email{vasiliev@spaziodati.eu}}

\maketitle              

\begin{abstract}
The most studied and most successful language models were developed and evaluated mainly for English and other close European languages, such as French, German, etc. It is important to study applicability of these models to other languages. The use of vector space models for Russian was recently studied for multiple corpora, such as Wikipedia, RuWac, lib.ru. These models were evaluated against word semantic similarity task. For our knowledge Twitter was not considered as a corpus for this task, with this work we fill the gap. Results for vectors trained on Twitter corpus are comparable in accuracy with other single-corpus trained models, although the best performance is currently achieved by combination of multiple corpora.
\end{abstract}
\section{Introduction}
Word semantic similarity task is an important part of contemporary NLP. It can be applied in many areas, like word sense disambiguation, information retrieval, information extraction and others. It has long history of improvements, starting with simple models, like bag-of-words (often weighted by TF-IDF score), continuing with more complex ones, like LSA \cite{LSA}, which attempts to find “latent” meanings of words and phrases, and even more abstract models, like NNLM \cite{NNLM}. Latest results are based on neural network experience, but are far more simple: various versions of Word2Vec, Skip-gram and CBOW models \cite{Word2Vec}, which currently show the State-of-the-Art results and have proven success with morphologically complex languages like Russian \cite{Arefyev}, \cite{Panchenko2015}. 

These are corpus-based approaches, where one computes or trains the model from a large corpus. They usually consider some word context, like in bag-of-words, where model is simple count of how often can some word be seen in context of a word being described. This model anyhow does not use semantic information. A step in semantic direction was made by LSA, which requires SVD transformation of co-occurrence matrix and produces vectors with latent, unknown structure. However, this method is rather computationally expensive, and can rarely be applied to large corpora. Distributed language model was proposed, where every word is initially assigned a random fixed-size vector. During training semantically close vectors (or close by means of context) become closer to each other; as matter of closeness the cosine similarity is usually chosen. This trick enables usage of neural networks and other machine learning techniques, which easily deal with fixed-size real vectors, instead of large and sparse co-occurrence vectors. 

It is worth mentioning non-corpus based techniques to estimate word semantic similarity. They usually make use of knowledge databases, like WordNet, Wikipedia, Wiktionary and others \cite{Zesch}, \cite{Bar}. It was shown that Wikipedia data can be used in graph-based methods \cite{Zesch2007}, and also in corpus-based ones. In this paper we are not focusing on non-corpus based techniques.

In this paper we concentrate on usage of Russian Twitter stream as training corpus for Word2Vec model in semantic similarity task, and show results comparable with current (trained on a single corpus). This research is part of molva.spb.ru project, which is a trending topic detection engine for Russian Twitter. Thus the choice of language of interest is narrowed down to only Russian, although there is strong intuition that one can achieve similar results with other languages.

\subsection{Goals of this paper}
The primary goal of this paper is to prove usefulness of Russian Twitter stream as word semantic similarity resource. Twitter is a popular social network\footnote{https://about.twitter.com/company}, or also called "microblogging service", which enables users to share and interact with short messages instantly and publicly (although private accounts are also available). Users all over the world generate hundreds of millions of tweets per day, all over the world, in many languages, generating enormous amount of verbal data.

Traditional corpora for the word semantic similarity task are News, Wikipedia, electronic libraries and others (e.g. RUSSE workshop \cite{Panchenko2015}). It was shown that type of corpus used for training affects the resulting accuracy. Twitter is not usually considered, and intuition behind this is that probably every-day language is too simple and too occasional to produce good results. On the other hand, the real-time nature of this user message stream seems promising, as it may reveal what certain word means in this given moment.  

The other counter-argument against Twitter-as-Dataset is the policy of Twitter, which disallows publication of any dump of Twitter messages larger than 50K \footnote{https://dev.twitter.com/overview/terms/policy}. However, this policy permits publication of Twitter IDs in any amount. Thus the secondary goal of this paper is to describe how to create this kind of dataset from scratch. We provide the sample of Twitter messages used, as well as set of Twitter IDs used during experiments \footnote{https://github.com/molva/molva/tree/master/dataset}.

\subsection{Previous work}
Semantic similarity and relatedness task received significant amount of attention. Several "Gold standard" datasets were produced to facilitate the evaluation of algorithms and models, including WordSim353 \cite{WordSim353}, RG-65 \cite{RG-65} for English language and others. These datasets consist of several pairs of words, where each pair receives a score from human annotators. The score represents the similarity between two words, from 0\% (not similar) to 100\% (identical meaning, words are synonyms). Usually these scores are filled out by a number of human annotators, for instance, 13 in case of WordSim353 \footnote{http://www.cs.technion.ac.il/~gabr/resources/data/wordsim353/wordsim353.html}. The inter-annotator agreement is measured and the mean value is put into dataset. 

Until recent days there was no such dataset for Russian language. To mitigate this the “RUSSE: The First Workshop on Russian Semantic Similarity”\cite{Panchenko2015} was conducted, producing RUSSE Human-Judgements evaluation dataset (we will refer to it as HJ-dataset). RUSSE dataset was constructed the following way. Firstly, datasets WordSim353, MC \cite{MC} and RG-65 were combined and translated. Then human judgements were obtained by crowdsourcing (using custom implementation). Final size of the dataset is 333 word pairs, it is available on-line\footnote{https://github.com/nlpub/russe-evaluation/blob/master/russe/evaluation/hj-test.csv}.

The RUSSE contest was followed by paper from its organizers \cite{Panchenko2015} and several participators \cite{Arefyev}, \cite{Kutuzov}, thus filling the gap in word semantic similarity task for Russian language. In this paper we evaluate a Word2Vec model, trained on Russian Twitter corpus against RUSSE HJ-dataset, and show results comparable to top results of other RUSSE competitors.
\section{Data processing}
In this section we describe how we receive data from Twitter, how we filter it and how we feed it to the model.
\subsection{Acquiring data}
Twitter provides well-documented API\footnote{https://dev.twitter.com/overview/documentation}, which allows to request any information about Tweets, users and their profiles, with respect to rate limits. There is special type of API, called Streaming API\footnote{https://dev.twitter.com/streaming/overview}, that provides a real-time stream of tweets. The key difference with regular API is that connection is kept alive as long as possible, and Tweets are sent in real-time to the client. There are three endpoints of Streaming API of our interest: “sample”, “filter” and “firehose”. The first one provides a sample (random subset) of the full Tweet stream. The second one allows to receive Tweets matching some search criteria: matching to one or more search keywords, produced by subset of users, or coming from certain geo location. The last one provides the full set of Tweets, although it is not available by default. In order to get Twitter “firehose” one can contact Twitter, or buy this stream from third-parties.

In our case the simplest approach would be to use “sample” endpoint, but it provides Tweets in all possible languages from all over the World, while we are concerned only about one language (Russian). In order to use this endpoint we implemented filtering based on language. The filter is simple: if Tweet does not contain a substring of 3 or more cyrillic symbols, it is considered non-Russian. Although this approach keeps Tweets in Mongolian, Ukrainian and other slavic languages (because they use cyrillic alphabet), the total amount of false-positives in this case is negligible. To demonstrate this we conducted simple experiment: on a random sample of 200 tweets only 5 were in a language different from Russian. In order not to rely on Twitter language detection, we chose to proceed with this method of language-based filtering.

However, the amount of Tweets received through “sample” endpoint was not satisfying. This is probably because “sample” endpoint always streams the same content to all its clients, and small portion of it comes in Russian language. In order to force mining of Tweets in Russian language, we chose "filter" endpoint, which requires some search query. We constructed heuristic query, containing some auxiliary words, specific to Russian language: conjunctions, pronouns, prepositions. The full list is as follows: 

\foreignlanguage{russian}{
я, у, к, в, по, на, ты, мы, до, на, она, он, и, да. }

We evaluated our search query on data obtained from “sample” endpoint, and 95\% of Tweets matched it. We consider this coverage as reasonable and now on use “filter” endpoint with the query and language filtering described above. In this paper we work with Tweet stream acquired from {\tt2015/07/21} till {\tt2015/08/04}. We refer to parts of the dataset by the day of acquisition: {\tt2015/07/21}, etc. Tweet IDs used in our experiments are listed on-line\footnote{https://github.com/molva/molva/tree/master/dataset}.
\subsection{Corpus preprocessing}
Corpus-based algorithms like BoW and Word2Vec require text to be tokenized, and sometimes to be stemmed as well. It is common practice to filter out Stop-Words (e.g. \cite{Kutuzov}), but in this work we don’t use it. Morphological richness of Russian language forces us to use stemming, even though models like Word2Vec does not require it. In our experiments stemmed version performs significantly better than unstemmed, so we only report results of stemmed one. To do stemming we use Yandex Tomita Parser \footnote{https://github.com/yandex/tomita-parser/}, which is an extractor of simple facts from text in Russian language. It is based on Yandex stemmer mystem\cite{Segalovich}. It requires a set of grammar rules and facts (i.e. simple data structures) to be extracted. In this paper we use it with one simple rule:
\begin{lstlisting}
S -> Word interp (SimpleFact.Word); 
\end{lstlisting}
This rule tells parser to interpret each word it sees and return it back immediately. We use Tomita Parser as we find it more user-friendly than mystem. Tomita Parser performs following operations: sentence splitting, tokenization, stemming, removing punctuation marks, transforming words to lowercase. Each Tweet is transformed into one or several lines of tab-separated sequences of words (if there are several sentences or lines in a Tweet).  Twitter-specific “Hashtags” and “User mentions” are treated by Tomita Parser as normal words, except that “@” and “\#” symbols are stripped off.

HJ-dataset contains non-lemmatized words. This is understandable, because the task of this dataset was oriented to human annotators. In several cases plural form is used (consider this pair: "\foreignlanguage{russian}{тигр}, \foreignlanguage{russian}{кошачьи}"). In order to compute similarity for those pairs, and having in mind that Twitter data is pre-stemmed, we have to stem HJ-dataset with same parser as well.   
\subsection{Training the model}
We use Word2Vec to obtain word vectors from Twitter corpus. In this model word vectors are initialized randomly for each unique word and are fed to a sort of neural network. Authors of Word2Vec propose two different models: Skip-gram and CBOW. The first one is trained to predict the context of the word given just the word vector itself. The second one is somewhat opposite: it is trained to predict the word vector given its context. In our study CBOW always performs worse than Skip-gram, hence we describe only results with Skip-gram model. Those models have several training parameters, namely: vector size, size of vocabulary (or minimal frequency of a word), context size, threshold of downsampling, amount of training epochs. We choose vector size based on size of corpus. We use “context size” as “number of tokens before or after current token”. In all experiments presented in this paper we use one training epoch. 

There are several implementations of Word2Vec available, including original C utility\footnote{https://code.google.com/p/word2vec/} and a Python library gensim\footnote{https://radimrehurek.com/gensim/models/word2vec.html}. We use the latter one as we find it more convenient. Output of Tomita Parser is fed directly line-by-line to the model. It produces the set of vectors, which we then query to obtain similarity between word vectors, in order to compute the correlation with HJ-dataset. To compute correlation we use Spearman coefficient, since it was used as accuracy measure in RUSSE \cite{Panchenko2015}.
\section{Experimental results}
In this section we describe properties of data obtained from Twitter, describe experiment protocols and results.
\subsection{Properties of the data}
In order to train Word2Vec model for semantic similarity task we collected Twitter messages for 15 full days, from {\tt2015/07/21} till {\tt2015/08/04}. Each day contains on average 3M of Tweets and 40M of tokens. All properties measured are shown in Table 1. Our first observation was that given one day of Twitter data we cannot estimate all of the words from HJ-dataset, because they appear too rarely. We fixed the frequency threshold on value of 40 occurrences per day and counted how many words from HJ-dataset are below this threshold. 

Our second observation was that words "missing" from HJ-dataset are different from day to day. This is not very surprising having in mind the dynamic nature of Twitter data. Thus estimation of word vectors is different from day to day. In order to estimate the fluctuation of this semantic measure, we conduct training of Word2Vec on each day in our corpus. We fix vector size to 300, context size to 5, downsampling threshold to 1e-3, and minimal word occurrence threshold (also called min-freq) to 40. The results are shown in Table 2. Mean Spearman correlation between daily Twitter splits and HJ-dataset is 0.36 with std.dev. of 0.04. Word pairs for missing words (infrequent ones) were excluded. We also create superset of all infrequent words, i.e. words having frequency below 40 in at least one daily split. This set contains 50 words and produces 76 "infrequent word" pairs (out of 333). Every pair containing at least one infrequent word was excluded. On that subset of HJ-dataset mean correlation is 0.29 with std.dev. of 0.03. We consider this to be reasonably stable result.

\begin{table}
\caption{Properties of Twitter corpus (15 full days)}
\begin{center}
\begin{tabular}{l r}
\hline
Type & Value\\[2pt]
\hline
\noalign{\vskip .1cm}    
Number of Tweets & 50M \\
Number of tokens & 580M \\
Number of sentences & 59M \\
Size of dictionary (full) & 13M \\
Size of dictionary (minfreq=40) & 236K \\
Number of tokens (minfreq=40) & 555M \\
Average sentence length & 9.80 \\
\hline
\end{tabular}
\end{center}
\end{table}
\begin{table}
\caption{Properties of Twitter corpus (average on daily slices)}
\begin{center}
\begin{tabular}{l r r}
\hline
Property & Mean & Std.Dev. \\
\hline
\noalign{\vskip .1cm} 
Number of Tweets & 3.3M  & 82.6K \\ 
Number of tokens & 38.6M & 0.985M \\
Missing pairs (infrequent)  & 44 & 5.44 \\
$R_{Pearson}$ with HJ-dataset (full set) & 0.36 & 0.04 \\
$R_{Pearson}$ with HJ-dataset (76 infrequent pairs filtered) & 0.29 & 0.03 \\
\hline
\end{tabular}
\end{center}
\end{table}
\subsection{Determining optimal corpus size}
Word2Vec model was designed to be trained on large corpora. There are results of training it in reasonable time with corpus size of 1 billion of tokens \cite{Word2Vec}. It was mentioned that accuracy of estimated word vectors improves with size of corpus. Twitter provides an enormous amount of data, thus it is a perfect job for Word2Vec. We fix parameters for the model with following values: vector size of 300, min-freq of 40, context size of 5 and downsampling of 1e-3. We train our model subsequently with 1, 7 and 15 days of Twitter data (each starting with {\tt07/21} and followed by subsequent days) . The largest corpus of 15 days contains 580M tokens. Results of training are shown in Table 3. In this experiment the best result belongs to 7-day corpus with 0.56 correlation with HJ-dataset, and 15-day corpus has a little less, 0.55. This can be explained by following: in order to achieve better results with Word2Vec one should increase both corpus and vector sizes. Indeed, training model with vector size of 600 on full Twitter corpus (15 days) shows the best result of 0.59. It is also worth noting that number of "missing" pairs is negligible in 7-days corpus: the only missing word (and pair) is "\foreignlanguage{russian}{йель}", Yale, the name of university in the USA. There are no "missing" words in 15-days corpus. 

\begin{table}
\caption{Properties of Twitter corpus (different size)}
\begin{center}
\begin{tabular}{l l l l l l}
\hline
$N_{days}$ & $N_{tokens}$ & Vector size & $R_{Spearman}$ & $R_{Pearson}$ \\[2pt]
\hline
\noalign{\vskip .1cm} 
1 & 38M & 300 & 0.39 & 0.39 \\
7 & 270M & 300 & 0.56 & 0.54 \\
10 & 580M & 300 & 0.55 & 0.54 \\
15 & 580M & 600 & \textbf{0.59} & 0.53 \\
\hline
\end{tabular}
\end{center}
\end{table}
Training the model on 15-days corpus took 8 hours on our machine with 2 cores and 4Gb of RAM. We have an intuition that further improvements are possible with larger corpus. Comparing our results to ones reported by RUSSE participants, we conclude that our best result of 0.598 is comparable to other results, as it (virtually) encloses the top-10 of results. However, best submission of RUSSE has huge gap in accuracy of 0.16, compared to our Twitter corpus. Having in mind that best results in RUSSE combine several corpora, it is reasonable to compare Twitter results to other single-corpus results. For convenience we replicate results for these corpora, originally presented in \cite{Panchenko2015}, alongside with our result in Table 5. Given these considerations we conclude that with size of Twitter corpus of 500M one can achieve reasonably good results on task of word semantic similarity.        
\subsection{Determining optimal context size}
Authors of Word2Vec \cite{Word2Vec} and Paragraph Vector \cite{Doc2Vec} advise to determine the optimal context size for each distinct training session. In our Twitter corpus average length of the sentence appears to be 9.8 with std.dev. of 4.9; it means that most of sentences have less than 20 tokens. This is one of peculiarities of Twitter data: Tweets are limited in size, hence sentences are short. Context size greater than 10 is redundant. We choose to train word vectors with 3 different context size values: 2, 5, 10. We make two rounds of training: first one, with Twitter data from days from {\tt07/21} till {\tt07/25}, and second, from {\tt07/26} till {\tt07/30}. Results of measuring correlation with HJ-dataset are shown in Table 4. According to these results context size of 5 is slightly better than others, but the difference is negligible compared to fluctuation between several attempts of training.
\begin{table}
\caption{$R_{Spearman}$ for different context size} 
\begin{center}
\begin{tabular}{l r r}
\hline
Date range & {\tt07/21-07/26} & {\tt07/26-07/30} \\
\hline
\noalign{\vskip .1cm} 
$C_{size} = 2$ & 0.3774 & 0.36166 \\
$C_{size} = 5$ & 0.3789 & 0.36859 \\
$C_{size} = 10$ & 0.3745 & 0.33798 \\
\hline
\end{tabular}
\end{center}
\end{table}
\begin{table}
\caption{Comparison with current single-corpus trained results}
\begin{center}
\begin{tabular}{l r r p{5cm}}
\hline
Corpus & Size, tokens & $R_{Spearman}$ & Method description \\
\hline
\noalign{\vskip .1cm} 
lib.rus.ec & 12B & 0.6537 & Word2vec trained on 150G of texts from lib.rus.ec (skip-gram, 500d vectors, window size 5, 3 iteration, min cnt 5) \\
RuWac & 2.0B & 0.6395 & GloVe (100d vectors) on RuWac (lemmatized, normalized) \\
\textbf{Twitter} & \textbf{580M} & \textbf{0.5987} & \textbf{Word2vec (skip-gram, window size 5, 600d vectors)} \\
\hline
\end{tabular}
\end{center}
\end{table}
\subsection{Some further observations}
Vector space model is capable to give more information than just measure of semantic distance of two given words. It was shown that word vectors can have multiple degrees of similarity. In particular, it is possible to model simple relations, like "country"-"capital city", gender, syntactic relations with algebraic operations over these vectors. Authors of \cite{Word2Vec} propose to assess quality of these vectors on task of exact prediction of these word relations. However, word vectors learned from Twitter seem to perform poorly on this task. We don’t make systematic research on this subject here because it goes outside of the scope of the current paper, though it is an important direction of future studies.

Twitter post often contains three special types of words: user mentions, hashtags and hyperlinks. It can be beneficial to filter them (consider as Stop-Words). In results presented in this paper, and in particular in Tables 3 and 4, we don’t filter such words. It is highly controversial if one should remove hashtags from analysis since they are often valid words or multiwords. It can also be beneficial, in some tasks, to estimate word vectors for a username. Hyperlinks in Twitter posts are mandatory shortened\footnote{An example of URL-shortener service: http://bit.ly}. It is not clear how to treat them: filter out completely, keep them or even un-short them. However, some of our experiments show that filtering of "User Mentions" and hyperlinks can improve accuracy on the word semantic relatedness task by 3-5\%. 
\section{Conclusion}
In this paper we investigated the use of Twitter corpus for training Word2Vec model for task of word semantic similarity. We described a method to obtain stream of Twitter messages and prepare them for training. We use HJ-dataset, which was created for RUSSE contest \cite{Panchenko2015} to measure correlation between similarity of word vectors and human judgements on word pairs similarity. We achieve results comparable with results obtained while training Word2Vec on traditional corpora, like Wikipedia and Web pages \cite{Arefyev}, \cite{Kutuzov}. This is especially important because Twitter data is highly dynamic, and traditional sources are mostly static (rarely change over time). Thus verbal data acquired from Twitter may be used to estimate word vectors for neologisms, or determine other changes in word semantic, as soon as they appear in human speech. 
%
%
\bibliographystyle{plain}

\begin{thebibliography}{10}

\bibitem{Arefyev}
Lukanin A. V. Lesota O.~O. Arefyev N.~V., Panchenko A.~I. and Romanov~P. V.
\newblock Evaluating three corpus-based semantic similarity systems for
  russian.
\newblock {\em Dialog 2015, Russia}, 2015.

\bibitem{Bar}
Daniel B{\"a}r, Chris Biemann, Iryna Gurevych, and Torsten Zesch.
\newblock Ukp: Computing semantic textual similarity by combining multiple
  content similarity measures.
\newblock In {\em Proceedings of the First Joint Conference on Lexical and
  Computational Semantics-Volume 1: Proceedings of the main conference and the
  shared task, and Volume 2: Proceedings of the Sixth International Workshop on
  Semantic Evaluation}, pages 435--440. Association for Computational
  Linguistics, 2012.

\bibitem{NNLM}
Yoshua Bengio, R{\'e}jean Ducharme, Pascal Vincent, and Christian Janvin.
\newblock A neural probabilistic language model.
\newblock {\em The Journal of Machine Learning Research}, 3:1137--1155, 2003.

\bibitem{WordSim353}
Lev Finkelstein, Evgeniy Gabrilovich, Yossi Matias, Ehud Rivlin, Zach Solan,
  Gadi Wolfman, and Eytan Ruppin.
\newblock Placing search in context: The concept revisited.
\newblock In {\em Proceedings of the 10th international conference on World
  Wide Web}, pages 406--414. ACM, 2001.

\bibitem{Kutuzov}
Andreev~I. Kutuzov~A.
\newblock Texts in, meaning out: Neural language models in semantic similarity
  tasks for russian.
\newblock {\em Dialog 2015, Russia}.

\bibitem{LSA}
Thomas~K Landauer, Peter~W Foltz, and Darrell Laham.
\newblock An introduction to latent semantic analysis.
\newblock {\em Discourse processes}, 25(2-3):259--284, 1998.

\bibitem{Doc2Vec}
Quoc~V Le and Tomas Mikolov.
\newblock Distributed representations of sentences and documents.
\newblock {\em arXiv preprint arXiv:1405.4053}, 2014.

\bibitem{Word2Vec}
Tomas Mikolov, Kai Chen, Greg Corrado, and Jeffrey Dean.
\newblock Efficient estimation of word representations in vector space.
\newblock {\em arXiv preprint arXiv:1301.3781}, 2013.

\bibitem{MC}
George~A Miller and Walter~G Charles.
\newblock Contextual correlates of semantic similarity.
\newblock {\em Language and cognitive processes}, 6(1):1--28, 1991.

\bibitem{Panchenko2015}
A.~Panchenko, N.~V. Loukachevitch, D.~Ustalov, D.~Paperno, C.~M. Meyer, and
  N.~Konstantinova.
\newblock {RUSSE: The First Workshop on Russian Semantic Similarity}.
\newblock In {\em Computational Linguistics and Intellectual Technologies:
  papers from the Annual conference ``Dialogue''}, volume~2, pages 89--105.
  RGGU, Moscow, 2015.

\bibitem{RG-65}
Herbert Rubenstein and John~B Goodenough.
\newblock Contextual correlates of synonymy.
\newblock {\em Communications of the ACM}, 8(10):627--633, 1965.

\bibitem{Segalovich}
Ilya Segalovich.
\newblock A fast morphological algorithm with unknown word guessing induced by
  a dictionary for a web search engine.
\newblock In {\em MLMTA}, pages 273--280. Citeseer, 2003.

\bibitem{Zesch2007}
Torsten Zesch and Iryna Gurevych.
\newblock Analysis of the wikipedia category graph for nlp applications.
\newblock In {\em Proceedings of the TextGraphs-2 Workshop (NAACL-HLT 2007)},
  pages 1--8, 2007.

\bibitem{Zesch}
Torsten Zesch, Christof M{\"u}ller, and Iryna Gurevych.
\newblock Using wiktionary for computing semantic relatedness.
\newblock In {\em AAAI}, volume~8, pages 861--866, 2008.

\end{thebibliography}

\end{document}